\documentclass[letterpaper,10pt,conference]{ieeeconf}

\pdfoutput=1 

\IEEEoverridecommandlockouts
\usepackage{amsmath}
\usepackage{amsfonts}
\usepackage{amssymb}

\usepackage[tight]{subfigure}
\usepackage[pdfusetitle,hidelinks]{hyperref}

\usepackage{epsfig}
\usepackage{graphicx}

\usepackage{multirow}

\usepackage{tabularx}
\usepackage{booktabs}

\usepackage{xspace} 

\usepackage[ruled]{algorithm2e}
\let\chapter\section 
\usepackage[]{algorithmic}

\usepackage{doi}

\usepackage[textsize=footnotesize]{todonotes}
\usepackage{flushend} 

\DeclareMathOperator*{\argmin}{arg\,min}

\makeatletter
\let\NAT@parse\undefined
\makeatother

\usepackage[numbers,sort&compress]{natbib}
\bibpunct{[}{]}{,}{n}{}{;}

\title{\LARGE \bf
Accurate Vision-based Vehicle Localization using Satellite Imagery
}

\author{{\normalfont\large Hang Chu \hspace{0.7cm} Hongyuan Mei \hspace{0.7cm} Mohit Bansal \hspace{0.7cm} Matthew R.~Walter}\\
  {\normalfont\normalsize Toyota Technological Institute at Chicago, Chicago, IL 60637, USA}\\
        {\tt\small \{hchu,hongyuan,mbansal,mwalter\}@ttic.edu}
}

\begin{document}

\maketitle

\begin{abstract}
    We propose a method for accurately localizing ground vehicles with
    the aid of satellite imagery. Our approach takes a ground image as
    input, and outputs the location from which it was taken on a
    georeferenced satellite image. We perform visual localization by
    estimating the co-occurrence probabilities between the ground and
    satellite images based on a ground-satellite feature
    dictionary. The method is able to estimate likelihoods over
    arbitrary locations without the need for a dense ground image
    database. We present a ranking-loss based algorithm that learns
    location-discriminative feature projection matrices that result in
    further improvements in accuracy. We evaluate our method on the
    Malaga and KITTI public datasets and
    demonstrate significant improvements over a baseline that performs
    exhaustive search.
\end{abstract}

\section{Introduction} \label{sec:intro}

Autonomous vehicles have recently received a great deal of attention
in the robotics, intelligent transportation, and artificial
intelligence communities. Accurate estimation of a vehicle's location
is a key capability to realizing autonomous operation. Currently, many
vehicles employ Global Positioning System (GPS) receivers to estimate
their absolute, georeferenced pose. However, most commercial GPS
systems suffer from limited precision and are sensitive to multipath
effects (e.g., in the so-called ``urban canyons'' formed by tall
buildings), which can introduce significant biases that are difficult
to detect. Visual place recognition seeks to overcome this limitation
by identifying a camera's (coarse) location in an a priori known
environment (typically in combination with map-based localization,
which uses visual recognition for loop-closure). Visual place
recognition, however, is challenging due to the appearance variations
that result from environment and perspective changes (e.g., parked
cars that are no longer present, illumination changes, weather
variations), and the perceptual aliasing that results from different
areas having similar appearance. A number of techniques have been
proposed of late that make significant progress towards overcoming
these challenges~\citep{cummins08, park10, cummins11, churchill12,
  milford12, johns13, sunderhauf13, naseer14, lynen14, mcmanus14,
  hansen14, sunderhauf15, sunderhauf15a}.

Satellite imagery provides an alternative source of information that
can be employed as a reference for vehicle localization. High
resolution, georeferenced, satellite images that densely cover the
world are becoming increasingly accessible and well-maintained, as
exemplified by the databases available via Google
Maps~\citep{google-maps} and the USGS~\cite{usgs}. Algorithms that are
able to reason over the correspondence between ground and satellite
imagery can exploit this availability to achieve wide-area camera
localization~\citep{jacobs07, bansal11, saticcvw11, lin13,
  viswanathan14, saticra15}.
\begin{figure}[!t]
    \centering
    \includegraphics[height=1.3in]{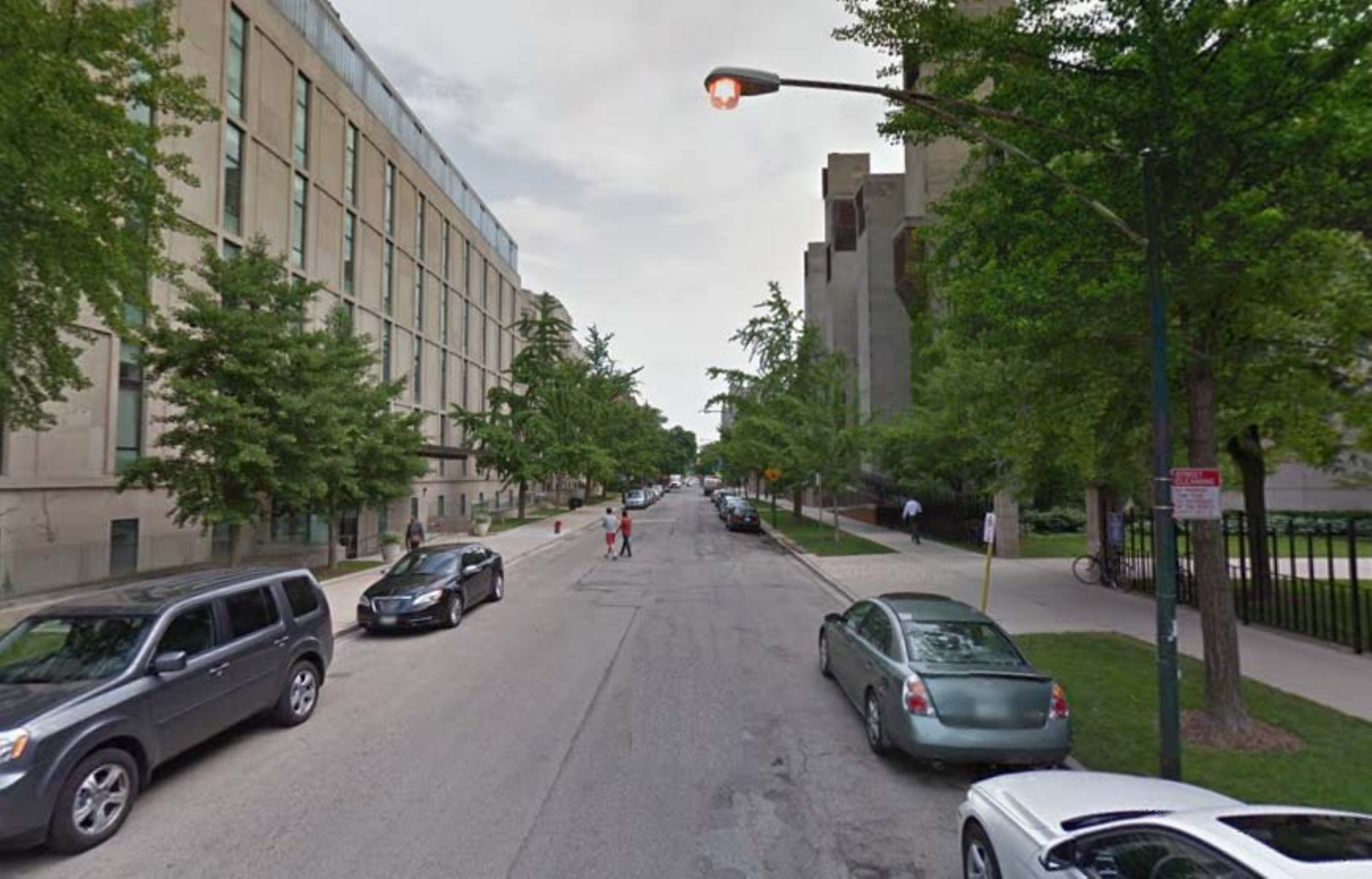}\hfil
    \includegraphics[height=1.3in]{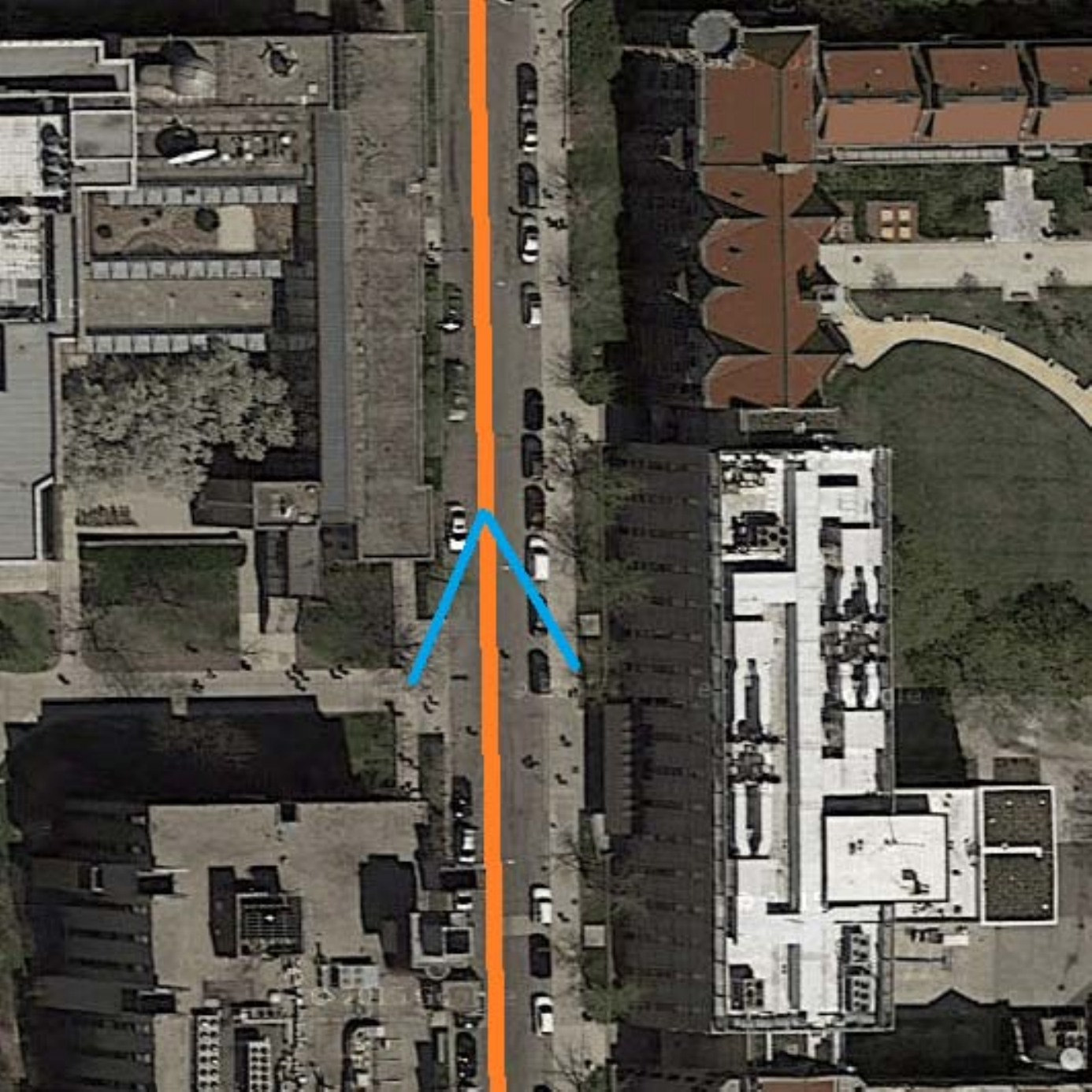}%
    \caption{Given a ground image (left), our method
      outputs the vehicle location (blue) on the satellite image
      (left), along the known vehicle path (orange).} \label{fig:ground-satellite-example}
\end{figure}

In this paper, we present a multi-view learning method that performs
accurate vision-based localization with the aid of satellite
imagery. Our system takes as input an outdoor stereo ground image and
returns the location of the stereo pair in a georeferenced satellite
image (Fig.~\ref{fig:ground-satellite-example}), assuming access to a
database of ground (stereo) and satellite images of the environment
(e.g., such as those acquired during a previous environment
visit). Instead of matching the query ground image against the
database of ground images, as is typically done for visual place
recognition, we estimate the co-occurrence probability of the query
ground image and the local satellite image at a particular location,
i.e., the probability over the ground viewpoint in the satellite
image. This allows us to identify a more precise distribution over
locations by interpolating the vehicle path with sampled local
satellite images.  In this way, our approach uses readily available
satellite images for localization, which improves accuracy without
requiring a dense database of ground images. Our method includes a
listwise ranking algorithm to learn effective feature projection
matrices that increase the features' discriminative power in terms of
location, and thus further improve localization accuracy.

The novel contributions of this paper are:
\begin{itemize}
\item We propose a strategy for localizing a camera based upon an
    estimate of the ground image-satellite image co-occurrence, which
    improves localization accuracy without ground image database
    expansion.
\item We describe a ranking-loss method that learns general feature
    projections to effectively capture the relationship between
    ground and satellite imagery.
\end{itemize}
%
%


\section{Related Work} \label{sec:related}

Several approaches exist that address the related problem of
localizing a ground image by matching it against an Internet-based
database of geotagged ground images. These methods typically employ visual
features~\citep{hays08,zheng09} or a combination of visual and
textual (i.e., image tags) features~\citep{crandall09}. These techniques
have proven effective at identifying the location of query images over
impressively large areas~\citep{hays08}. However, their reliance
upon available reference images limits their use to regions
with sufficient coverage (e.g., those visited by tourists) and their
accuracy depends on the spatial density of this coverage. Meanwhile,
several methods~\citep{schindler07,zamir10,chen11a,majdik13} treat
vision-based localization as a problem of image retrieval against a
database of street view images. \citet{majdik13} propose a method that
localizes a micro aerial vehicle within urban scenes by matching
against virtual views generated from a Google Street View
image database.

In similar fashion to our work, previous methods have investigated
visual localization of ground images against a collection of
georeferenced satellite
images~\citep{jacobs07,bansal11,lin13,viswanathan14}. \citet{bansal11}
describe a method that localizes street view images relative to a
collection of geotagged oblique aerial and satellite images. Their
method uses the combination of satellite and aerial images to extract
building facades and their locations. Localization then follows by
matching these facades against those in the query ground
image. Meanwhile, \citet{lin13} leverage the availability of land use
attributes and propose a cross-view learning approach that learns the
correspondence between features from ground-level images, overhead
images, and land cover data. \citet{viswanathan14} describe an
algorithm that warps panoramic ground images to obtain a projected bird's eye
view of the ground that they then match to a grid of satellite
locations. The inferred poses are then used as observations in a
particle filter for tracking. Other
methods~\citep{saticra15,saticcvw11} focus on extracting
orthographical texture patterns (e.g., road lane markings on the
ground plane) and then match these observed patterns with the
satellite image. These approaches perform well, but rely on the
existence of clear, non-occluded visual textures. Meanwhile, other
work has considered the related task of of visual localization of a
ground robot relative to images acquired with an aerial
vehicle~\citep{stentz02,kelly06}.

A great deal of attention has been paid in the robotics and vision
communities to the related problem of visual place recognition~\citep{cummins08,
  park10, cummins11, churchill12, milford12, johns13, sunderhauf13,
  naseer14, lynen14, mcmanus14, hansen14, sunderhauf15, sunderhauf15a}. The biggest
challenges to visual place recognition arise due to variations in image
appearance that result from changes in viewpoint, environment
structure, and illumination, as well as to perceptual aliasing, which are
both typical of real-world environments. Much of the work seeks to mitigate some of these challenges by using
interest point detectors and descriptors that are robust to transformations in scale and
rotation, as well as to slight variations in illumination (e.g., SIFT~\cite{sift} and
SURF~\cite{surf}).  Place
recognition then follows as image retrieval, i.e., image-to-image
matching-based search against a database (with various methods to improve
efficiency)~\citep{wolf05,li06,filliat07,schindler07}. While these
methods have demonstrated reasonable performance despite some appearance
variations, they are prone to failure when the environment is
perceptually aliased. Under these conditions, feature descriptors are
no longer sufficiently discriminative, which results in false matches
(notably, when the query image corresponds to a environment location
that is not in the map). When used for data association in a
downstream SLAM framework, these erroneous loop closures can result in
estimator divergence.

\begin{figure*}[!t]
    \centering
    \includegraphics[width=0.87\textwidth]{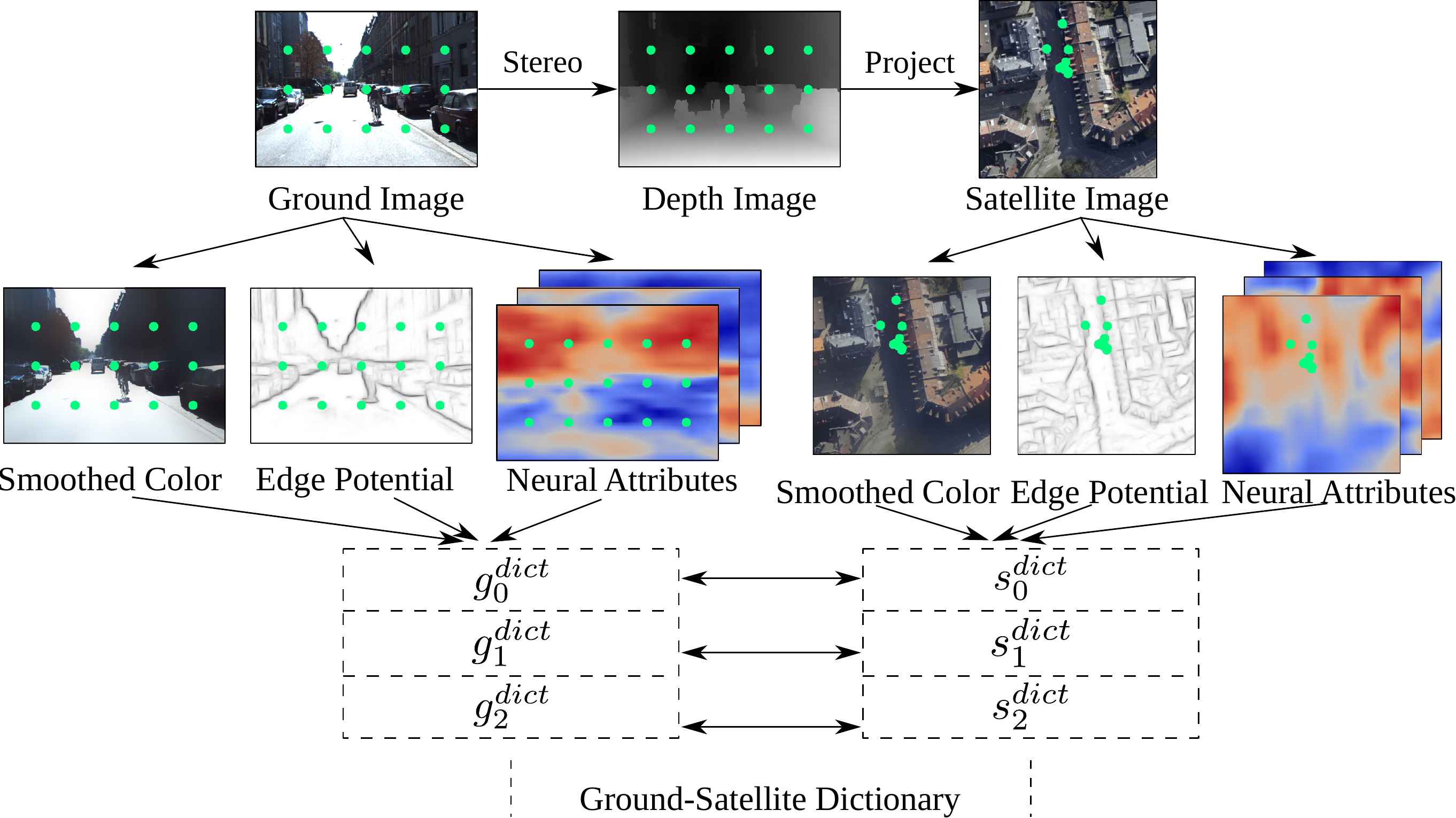}
    \caption{Illustration of our ground and satellite feature dictionary learning process.}
    \label{fig:dictionary}
\end{figure*}
The FAB-MAP algorithm by Cummins and Newman \cite{cummins08,cummins11} is
designed to address challenges that arise as a result of perceptual
aliasing. To do so, FAB-MAP learns a generative model of region
appearance using a bag-of-words representation that expresses the
commonality of certain features. By essentially modeling this
perceptual ambiguity, the authors are able to reject invalid image
matches despite significant aliasing, while correctly recognizing
those that are valid. Alternatively, other methods achieve
robustness to perceptual aliasing and appearance variations by
treating visual place recognition as a problem of matching image
sequences~\citep{koch10,milford12,johns13,sunderhauf13,naseer14},
whereby imposing joint consistency reduces the likelihood of false
matches. The robustness of image retrieval methods can be further
improved by increasing the space of appearance variations spanned by
the database~\citep{neubert13,mcmanus14a}. However, approaches that
achieve invariance proportional to the richness of their training
necessarily require larger databases to achieve robustness. Our method
similarly learns a probabilistic model that we can then use for
matching, though our likelihood model is over
the location of the query in the georeferenced satellite
image. Rather than treating localization as image retrieval, whereby we find
the nearest map image for a given query, we instead leverage the
availability of satellite images to estimate the interpolated
position. This provides additional robustness to viewpoint variation,
particularly as a result of increased separation between the
database images (e.g., on the order of 10\,m).

Meanwhile, recent attention in visual place recognition has focused on
the particularly challenging problem of identifying matches in the
event that there are significant appearance variations due to large
illumination changes (e.g., matching a query image taken at night to a
database image taken during the day) and seasonal changes (e.g.,
matching a query image with snow to one taken during
summer)~\citep{neubert13,mcmanus14,maddern14,lowry14,sunderhauf15,sunderhauf15a}. While
some of these variations can be captured with a sufficiently rich
database~\citep{neubert13}, this comes at the cost of requiring a
great deal of training data and its ability to generalize is not
clear~\citep{sunderhauf15}. \citet{mcmanus14a} seek to overcome the
brittleness of point-based features to environment
variation~\citep{valgren07, glover10, milford12, naseer14} by learning
a region-based detector to improve invariance to
appearance changes. Alternatively, \citet{sunderhauf15}  build on the
recent success of deep neural networks and eschew
traditional features in favor of ones that can be learned from
large corpora of images~\citep{russakovski14}. Their framework first
detects candidate landmarks in an image using state-of-the-art
proposal methods, and then employs a convolutional neural network to generate
features for each landmark. These features provide robustness to
appearance and viewpoint variations that enables accurate place
recognition under challenging environmental conditions. We also take advantage of this recent
development in convolutional neural networks for image
segmentation to achieve effective pixel-wise semantic
feature extractors.  

Related to our approach of projection matrix learning is a long
history of machine learning research with ranking loss objective
functions~\cite{hang11}. Our approach is also related to the field of
multi-view learning~\cite{xu13}, as we learn the relationship between ground
and satellite image views.


\section{Approach} \label{sec:approach}

Our approach first constructs a ground-satellite feature
dictionary that captures the relationship between the two views using
an existing database from the area. Second, we learn projection
matrices for the each of the two views so as to arrive at a feature
embedding that is more location-discriminative. Given a query ground
image, we then use the dictionary and the learned projections to
arrive at a distribution over the location in georeferenced satellite
images of the environment in which
the ground image was taken. Next, we describe these three steps in detail.

\subsection{Ground-Satellite Image Feature Dictionary}

The role of the ground-satellite dictionary is to capture the
relationships that exists between feature-based representations of
ground images and their corresponding overhead view. Specifically, we
define our ground-satellite image dictionary using three types of
features. The first consists of pixel-level RGB intensity, which we
smooth using bilateral filtering to preserve color
transitions~\cite{fastbilateral}. The second takes the form of
edge potentials for which we use a structured forest-based
method~\cite{edgedet} that is more robust to non-semantic noise (as
compared to classic gradient-based edge detectors) and can be computed
in real-time. The third feature type we use are neural, pixel-wise,
dense, semantic attributes. For these, we use fully-convolutional
neural networks~\cite{fcn} trained on ImageNet~\cite{imagenet} and
fine-tuned on PASCAL VOC~\cite{pascalvoc2012}. Next, we describe the process by which we construct the
ground-satellite dictionary, which we depict in Figure~\ref{fig:dictionary}.

For each ground image in the database, we identify the corresponding
satellite image centered on and oriented with the ground image
pose. We then compute pixel-wise features for both images. Next, we
compute the ground image features $g^{dict}_{i}$ on a
fixed-interval 2D grid (to ensure an efficient dictionary size), and
project these sampled points onto the satellite image using the
stereo-based depth estimate~\cite{stereo}.\footnote{For the
  experimental evaluation, we use the stereo pair only to estimate
  depth. We only use images from one camera to generate the
  dictionary, learn the projections, and estimate pose.} Sampled points that fall
outside the satellite image region are rejected. We record the
satellite features corresponding to the remaining projected points,
which we denote as $s^{dict}_{i}$. We repeat this process for all ground-satellite image
pairs to form our one-to-one ground-satellite image feature
dictionary. We also store dense satellite features even if they do not
appear in the dictionary, so that they do not need to be recomputed
during the localization phase. We store the dictionary with two $k$-d
trees for fast retrieval.

\subsection{Location-Discriminative Projection Learning}

The goal of location-discriminative projection learning is to identify
two linear projections $W_g$ and $W_s$ that transform the ground and
satellite image features
such that those that are physically close are also nearby in the
projected space. Nominally,
we can
model this projection learning task as
an optimization over a loss function that expresses the distance
in location between each feature point and its nearest neighbor
in the projected space. In the case of the ground image, this results
in a loss of the form
\begin{equation}
  W_g=\argmin_{W}\sum\limits_{i}\Delta L\bigl(i,\argmin_{k\in \mathcal{N}(i)}f_g(i,k,W)\bigr),
  \label{eqn:cost}
\end{equation}
where $\Delta L(i,k)$ is the
scalar difference in location (ignoring orientation for simplicity) between two feature
points, $\mathcal{N}(i)$ is a neighborhood around feature $i$ in
feature space, and
\mbox{$f_g(i,k,W) = \lVert Wg^{dict}_{i}-Wg^{dict}_{k}\rVert_{2}$}. We
consider a feature-space neighborhood for computational efficiency and
have found a $\mathcal{N}(i)=20$ neighborhood to be
effective in our reported experiments. A similar definition is used for $W_s$.

In practice, however, this loss function is difficult to
optimize. Instead, we treat the objective as a ranking problem and optimize
over a surrogate loss function that employs the hinge loss,
\begin{equation}
        \mathcal{L} \!=\! \sum\limits_{i}\!\left(\!f_g(i,k^*,W) \!-\!\!\! \min\limits_{k\in \mathcal{N}(i)}\!\!\Big(f_g(i,k,W)\!-\!m(i,k)\Big)\!\right)_{+}\!\!\!\!
        \label{eqn:hingeloss}
\end{equation}
where \mbox{$k^* = \argmin_{k}\Delta L(i,k)$} is the nearest feature
with regards to metric location, \mbox{$m(i,k) = \Delta L(i,k) - \Delta L(i,k^*)$}, and
\mbox{$(x)_+ = \textrm{max}(0,x)$} denotes the hinge loss.  Intuitively, we would like
the distance $f_g(i,k^*,W)$ to be smaller than any other distance $f_g(i,k,W)$ by a margin
\mbox{$m(i,k)$}.  We minimize the loss function~\eqref{eqn:hingeloss}
using stochastic gradient descent with Adam~\cite{adam} as the weight
update algorithm. Algorithm~\ref{algo:proj} describes the process we
use to learn a projection matrix, which is repeated twice for both $W_g$,
and $W_s$.
\begin{algorithm}[!t]
    \caption{Location-discriminative projection learning\!}
    \textbf{Input:} $\{g^{dict}_i\}$, $\{L(i)\}$\\
    \textbf{Output:} $W$\\
    \begin{algorithmic}[1]
        \STATE Initialize $W=\mathbb{I}$ and $t=0$
        \FOR{$epi$=1:\textsc{MaxIter}}
			\FOR{each $i$}
				\STATE $t = epi \times i_{max} + i$
				\IF{$ f_g(i,k^*\!,W) \!-\!\!\! \min\limits_{k\in \mathcal{N}(i)}\!\!\Big(\!f_g(i,k,W)\!-\!m(i,k)\!\Big) \!>\! 0$}
					\STATE Compute $\partial \mathcal{L}_t$ as \mbox{$ \partial \Big(f_g(i,k^*,W) - \min\limits_{k\in \mathcal{N}(i)}f_g(i,k,W) \Big) / \partial W$}
					\STATE Compute $\Delta W_t$ as \mbox{$\textsc{Adam}(\{\partial \mathcal{L}_0, \partial \mathcal{L}_1, \ldots, \partial \mathcal{L}_t \})$ }~\cite{adam}
					\STATE Update $W \leftarrow W - \Delta W_t$
				\ENDIF
			\ENDFOR
			\IF{convergence}
				\STATE break
			\ENDIF
        \ENDFOR
    \end{algorithmic}
    \label{algo:proj}
\end{algorithm}

\subsection{Localization}
\begin{figure}[!t]
    \centering
    \subfigure[]{\includegraphics[width=0.45\linewidth]{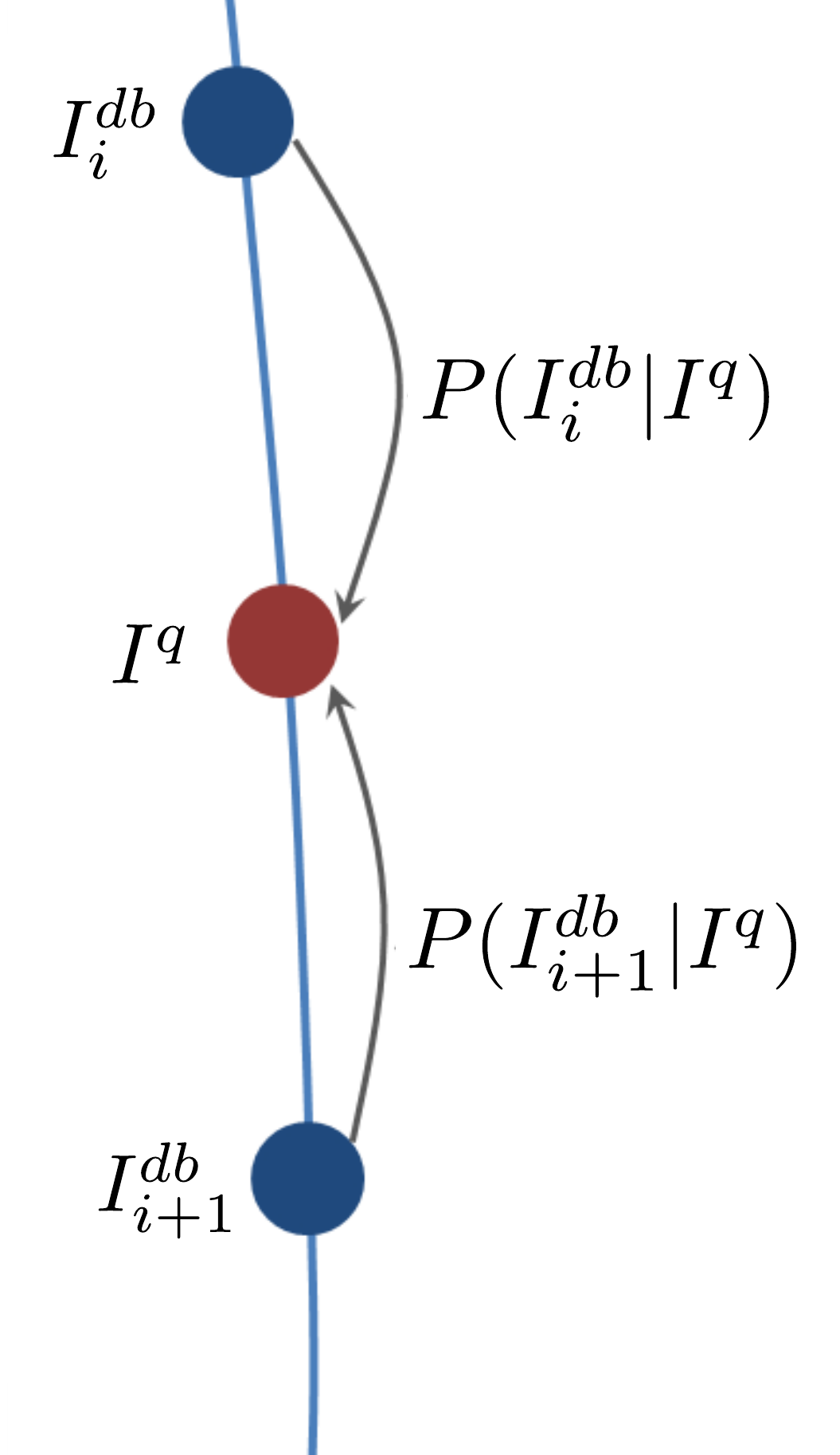}} \hfil%
    \subfigure[]{\includegraphics[width=0.45\linewidth]{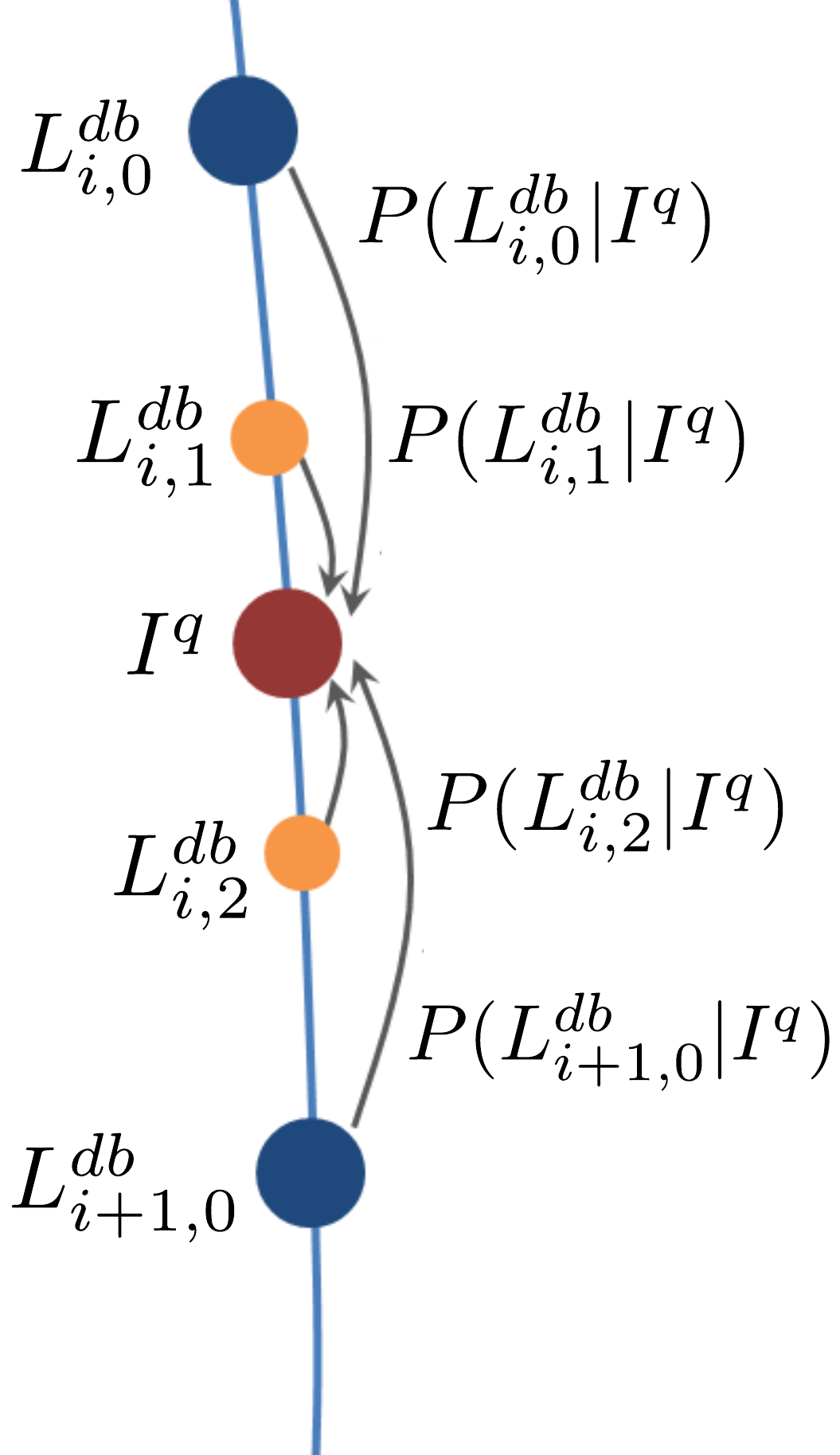}}
    \caption{Localization by (a) image-to-image matching using
      two-image reference database. By estimating the
      ground-satellite co-occurrence, our method (b) yields a more fine-grained distribution over the camera's location, without the
      need for a larger ground image database. $I_q$ denotes the query
      ground image. $I_i^{db}$ denotes a database ground image, and
      $L_{i,0}^{db}$. $L_{i,1}^{db}$ and $L_{i,2}^{db}$
      are the interpolated locations along the vehicle path. Our
      method is able to evaluate localization possibilities of
      $L_{i,1}^{db}$ and $L_{i,2}^{db}$ without knowing their ground
      images.}
    \label{fig:interpolation}
\end{figure}
In our final localization step, we compute the probability
$P(L|I^{q})$ that a given query ground image $I^q$ was taken at a
particular position and orientation $L$, where we interpolate the
database locations along the vehicle path to get a larger number of
location candidates $L$ (Fig.~\ref{fig:interpolation}). In order to arrive at
this likelihood, we first extract features for the query image $I^q$
and then retrieve the precomputed dense features for the satellite
image associated with pose $L$. Next, we sample the query ground image
features with a 2D grid and find their corresponding satellite image
features by projecting using the query stereo image, as if $I^{q}$ was
centered and oriented at pose $L$. After rejecting projected samples that
lie outside the satellite image, we obtain a set of ground-satellite
image feature pairs, where the $n^{th}$ pair is denoted as
$(g^{q}_{n},s^{L}_{n})$.

For each ground-satellite image feature pair, we evaluate their
co-occurrence score according to the size of the intersection of their
respective database neighbor sets in the projected feature space. To
do so, we first transform the features to their corresponding projected
spaces as $W_gg^{q}_{n}$ and
$W_ss^{L}_{n}$. Next, we retrieve their $M$-nearest neighbors, each for
the transformed ground and satellite images, among the dictionary
features projected into the projected space using the Approximate Nearest Neighbor
algorithm~\cite{ann}. The retrieved neighbor index sets are denoted as
$\{id_g^m(W_gg^{q}_{n})\}$ and $\{id_s^m(W_ss^{L}_{n})\}$. The
Euclidean feature distances between the query and database index $m$
are denoted as $\{d_g^m(W_gg^{q}_{n})\}$ and $\{d_s^m(W_ss^{L}_{n})\}$
for the ground and satellite images, respectively.  A single-pair
co-occurrence score is expressed as the consistency between the two
retrieved sets
\begin{equation}
  \textrm{score}(s^{L}_{n}|g^{q}_{n})=\!\!\!\!\sum\limits_{(m_1,m_2)\in
    I}\!\!\!\!\Big(d_g^{m_1}(W_gg^{q}_{n})\cdot d_s^{m_2}(W_ss^{L}_{n})\Big)^{-1},
\end{equation}
where $I=\{id_g^m(W_gg^{q}_{n})\}\cap \{id_s^m(W_ss^{L}_{n})\}$ denotes
all the $(m_1,m_2)$ pairs that are in the intersection of the two sets.
We then define the desired probability over the pose $L$ for the
query image $I^{q}$ as
\begin{equation}
  P(L|I^{q})=\frac{1}{C}\sum\limits_{n}\textrm{score}(s^{L}_{n}|g^{q}_{n}),
\end{equation}
where $C$ is a normalizing factor. We interpolate the database
vehicle path with $L_{i,j}^{db}$ (Fig.~\ref{fig:interpolation})
and infer the final location as that where $P(L|I^{q})$ is
maximized.


\section{Experimental Results} \label{sec:results}

We evaluate our method on the widely-used, publicly available
KITTI~\cite{kitti} and Malaga-Urban~\cite{malaga} datasets.

\subsection{KITTI Dataset}

\begin{figure}[!t]
    \centering
    \includegraphics[width=0.475\linewidth]{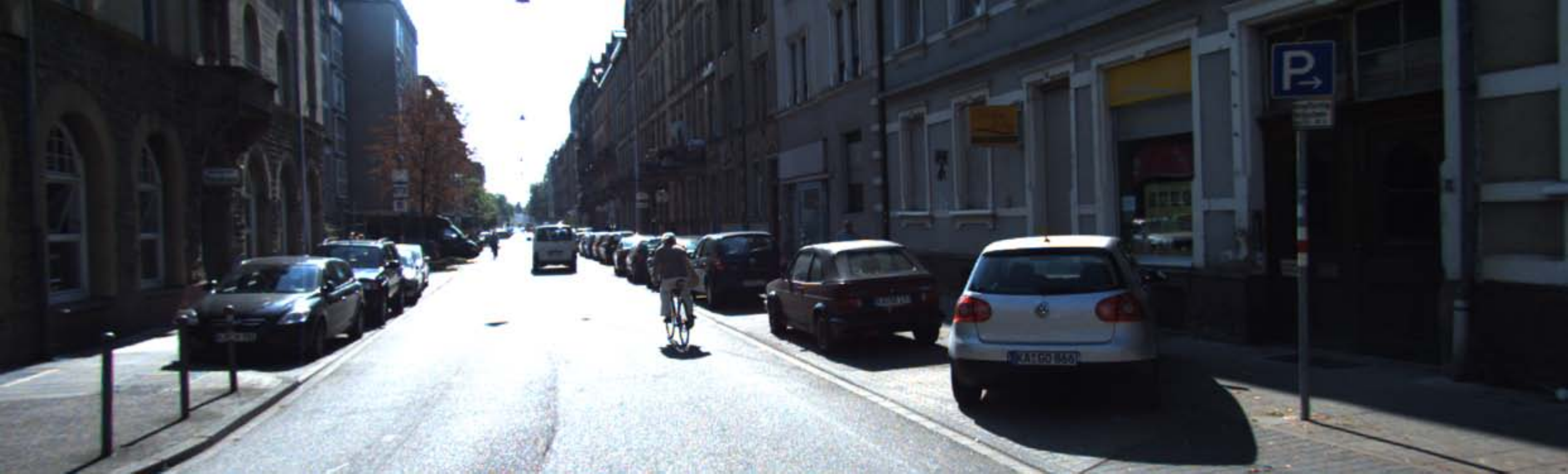}\hfil
    \includegraphics[width=0.475\linewidth]{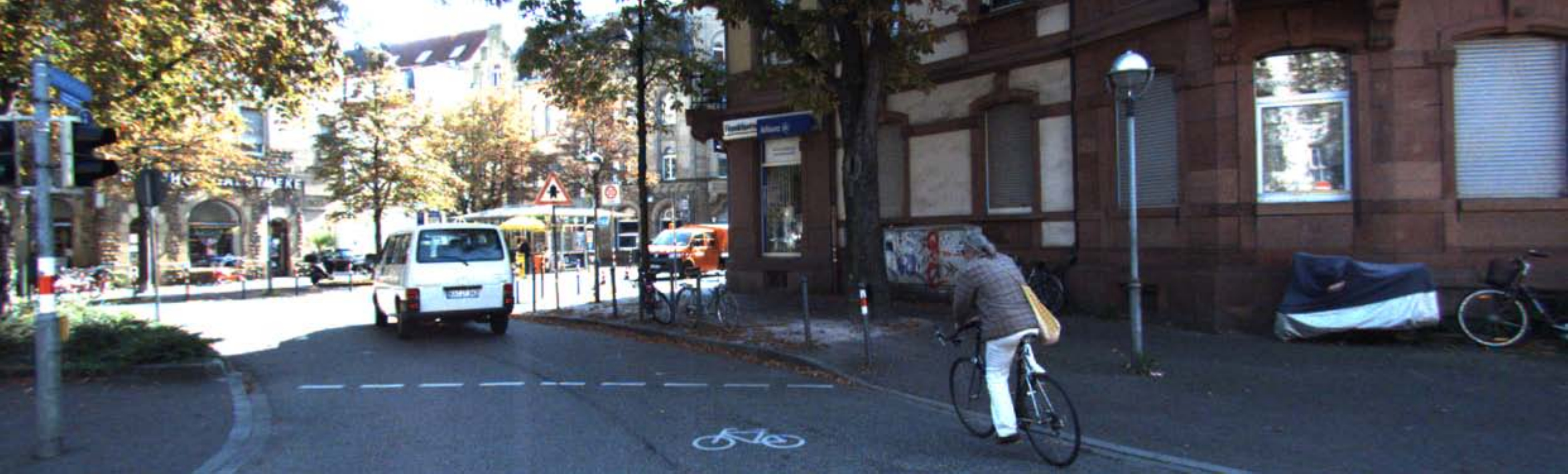}\\
    \vspace{0.25cm}
    \includegraphics[height=2.0in]{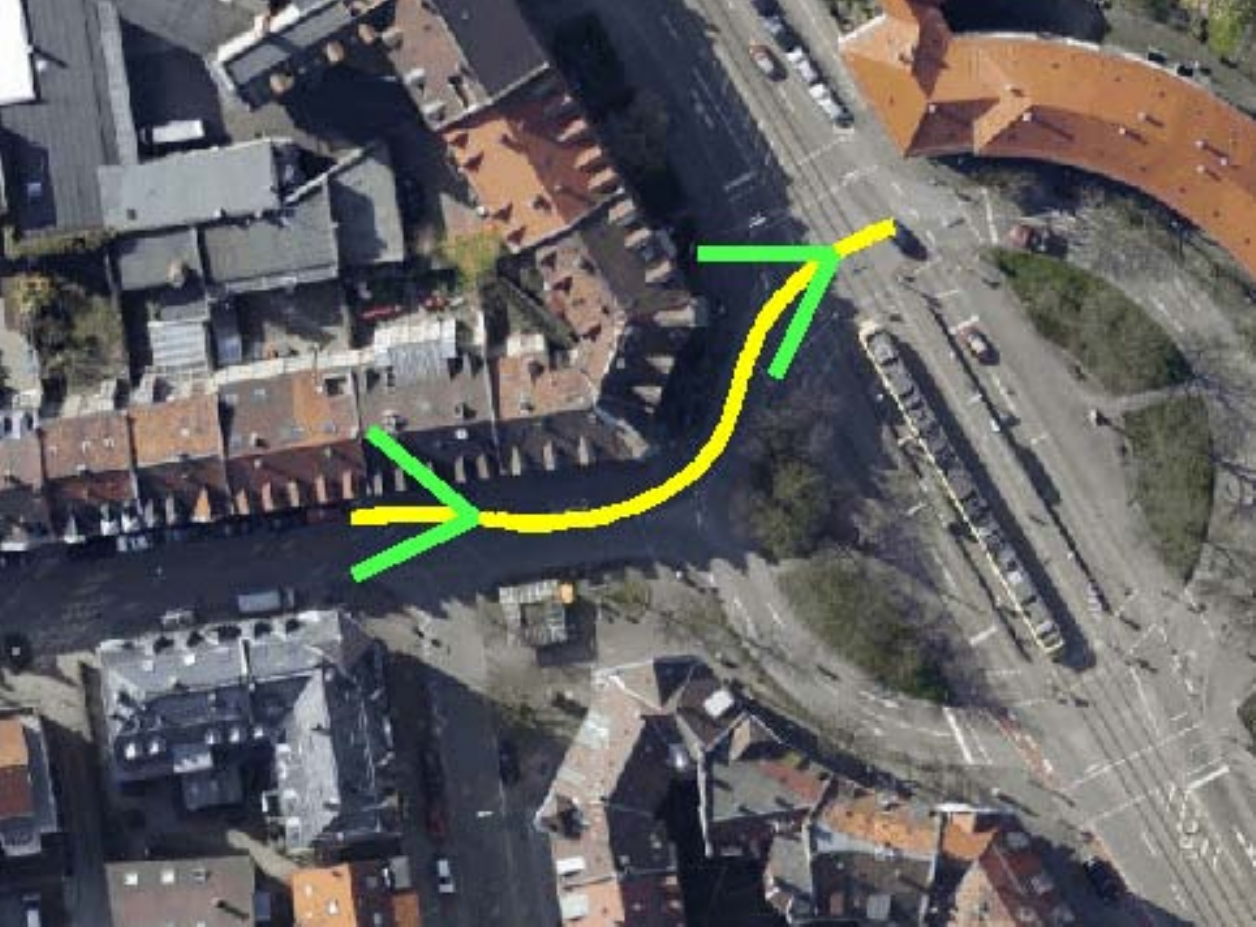}
    \caption{Examples from the KITTI-City dataset that include (top)
      two ground images and (bottom) a satellite image with a curve
      denoting the vehicle path and arrows that indicate the poses
      from which the ground images were taken.}
    \label{fig:kitti-city}
\end{figure}
\begin{figure}[!t]
    \centering
    \includegraphics[width=0.475\linewidth]{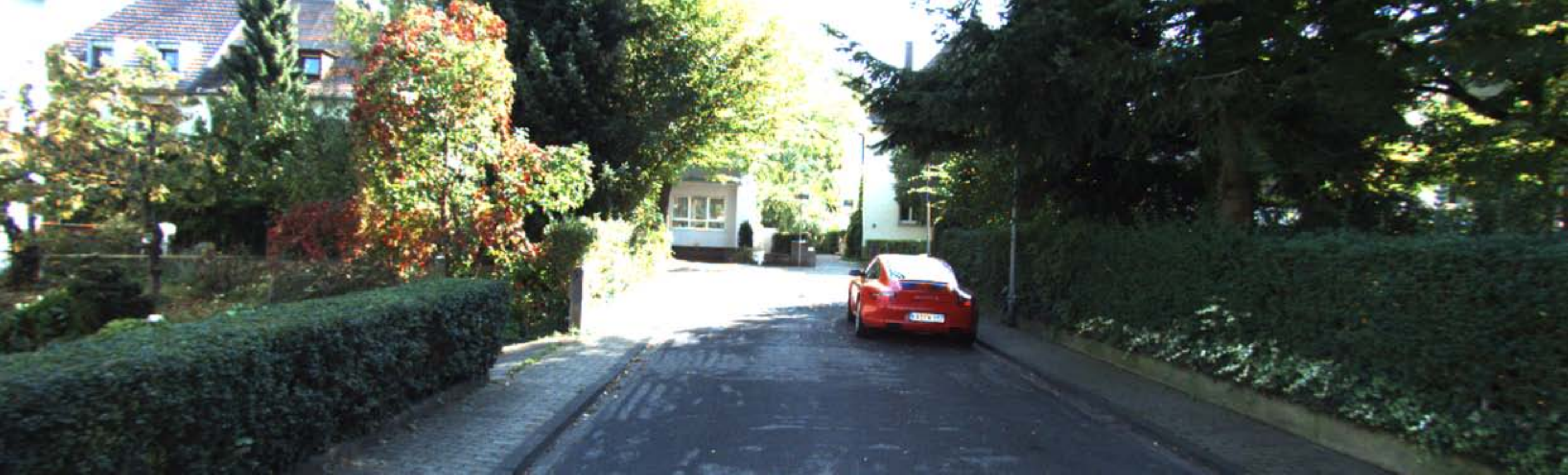}\hfil
    \includegraphics[width=0.475\linewidth]{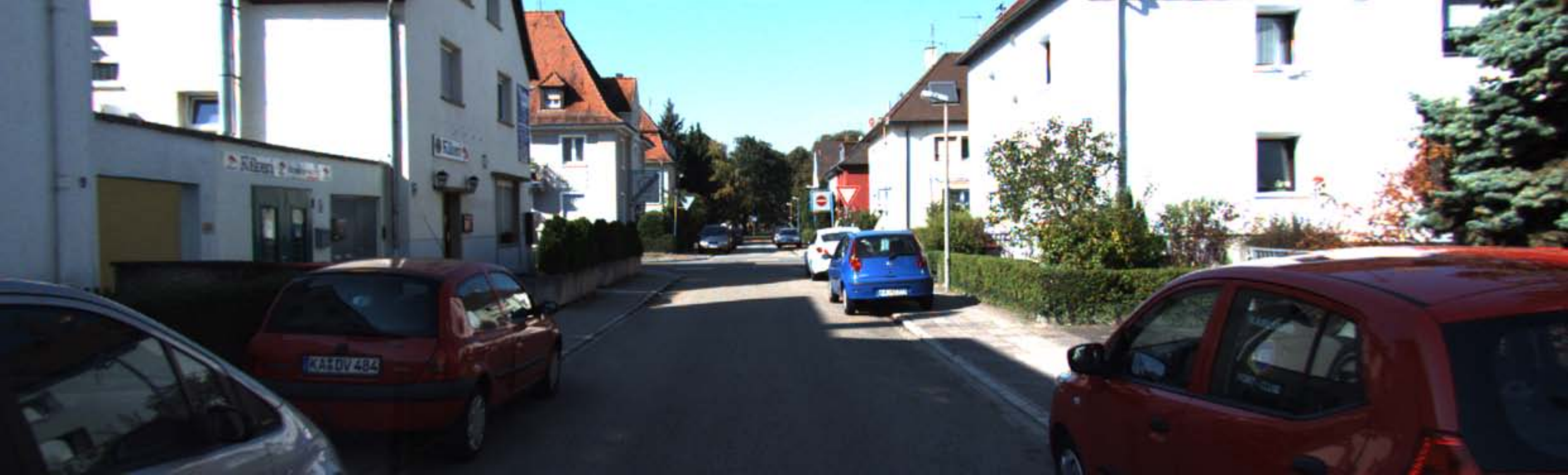}%
    \vspace{0.25cm}
    \includegraphics[height=2.0in]{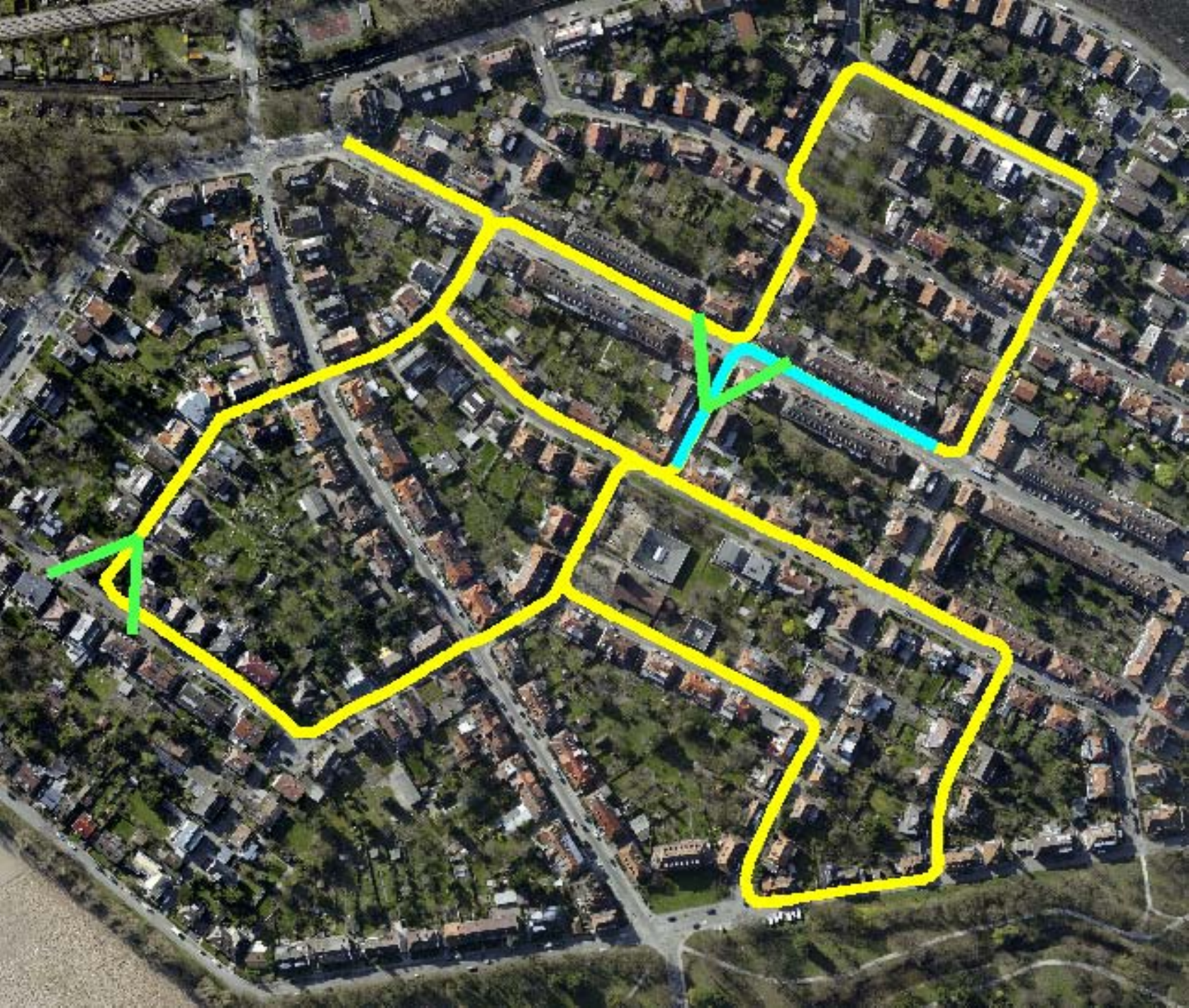}%
    \caption{Examples from the KITTI-Residential dataset that include
      (top) two ground images and (bottom) a satellite image. The
      yellow and cyan paths denote the vehicle path during the first and second visit, respectively. Green arrows
      indicate the pose from which the images were taken.}
\label{fig:kitty-residential}
\end{figure}

We conduct two experiments on the KITTI dataset. In the first
experiment, we use five raw data sequences from the KITTI-City
category. Together, these sequences involve $1067$ total images,
each with a resolution of $1242\times 375$. The total driving distance
for these sequences is $822.9$\,m. We randomly select $40\%$ of the
images as the database image set, and the rest as the query image
set. Figure~\ref{fig:kitti-city} provides examples of these images and the
vehicle's path. In the second KITTI experiment, we consider a scenario in which the
vehicle initially navigates an environment and later uses the
resulting learned database for localization when revisiting the area.
We emulate this scenario using a long sequence drawn from
the KITTI-Residential category, where we use the ground images from
the first pass through the environment to generate the database, and
images from the second visit as the query set. For the long
KITTI-Residential sequences, we downsample the
database and query sets at a rate of approximately one image per second
to reduce the overlap in viewpoints, which facilitates
  the independence assumptions of FAB-MAP\citep{cummins11}. This
results in $3654$ total images ($3376$ images for the database and
$278$ images for the query), each with a resolution of
$1241\times 376$, with a total of $3080$\,m travelled by the vehicle
($2847$\,m and $233$\,m for the database and query,
respectively). Figure~\ref{fig:kitty-residential} shows example ground
images and the database/query data split. Note that while the
ground-satellite dictionary is generated using images from the
downsampled database, projection learning was performed using the
full-framerate imagery along the database path.

We evaluate three variations of our method on these datasets as a
means of ablating individual components, and compare against two
existing approaches, which serve as baselines. One baseline that we
consider is a publicly available implementation of
FAB-MAP~\cite{cummins11},\footnote{\url{https://github.com/arrenglover/openfabmap}}
for which we use SURF features with a
cluster-size of $0.45$, which yields $2611$ and $3800$ bag-of-words
for the two experiments, respectively. We consider a second baseline
that performs exhaustive matching over a dense set of SURF features
for image retrieval. We further refine these matches using
RANSAC-based~\citep{fischler81} image-to-image homography estimation
to identify a set of geometrically consistent inlier features. We use
the average feature distance over these inliers as the final
measurement of image-to-image similarity. We refer to this baseline as
Exhaustive Feature Matching (EFM). The first variation of our method
that we consider eschews satellite images and instead performs ground
image retrieval using our image features (as opposed to SURF), which
we refer to as Ours-Ground-Only (Ours-GO). The second ablation
consists of our proposed framework with satellite images, but without
learning the location-discriminative feature projection, which we
refer to as Ours-No-Projection (Ours-NP). Lastly, we consider our full
method that utilizes satellite images and all images along the
database path for projection learning. For all methods,
  we identify the final position as a weighted average of the top
  three location matches. We also considered an
experiment in which we include query images taken from regions
outside those captured in our learned database and found that none of
the methods produced any false positives, which we define as
location estimates that are more than $10$\,m from ground-truth.

\begin{table}[!th]
    \centering
    \caption{KITTI localization error (meters)}
    \label{tab:kitti}
    \begin{tabularx}{0.75\linewidth}{lcc}
      \toprule
      Method                   &  KITTI-City         & KITTI-Residential\\
      \midrule
      FAB-MAP \cite{cummins11} & 1.24 (0.69)          & 2.29 (1.55)         \\
      EFM                      & 0.87 (0.15)          & 1.18 (0.91)         \\
      Ours-GO                  & 0.81 (0.07)          & 1.13 (0.81)         \\
      Ours-NP                  & 0.41 (0.20)          & 0.62 (0.33)         \\
      Ours-full                & \textbf{0.39} (0.22) & \textbf{0.42} (0.20)\\ 
      \bottomrule
    \end{tabularx}
\end{table}
Table~\ref{tab:kitti} compares the localization error for each of the
methods on the two KITTI-based experiments, with our method
outperforming the FAB-MAP and EFM baselines. Ours-GO achieves lower
error than the two SURF-based methods, which shows the effectiveness
of our proposed features at discriminating between ground images,
especially when there is overlap between images.\footnote{We note that
  this may violate independence assumptions that
  are made when learning the generative model for FAB-MAP~\citep{cummins11}.}
Ours-NP further reduces the error by interpolating the trajectory
between two adjacent ground database images (as described in
Fig.~\ref{fig:interpolation}) and evaluating ground-satellite
co-occurrence probabilities, which brings in more localization
information. Ours-full achieves the lowest error, which demonstrates
the effectiveness of using re-ranking to learn the
location-discriminative projection matrices. Note that Ours-NP and
Ours-full use stereo to compute depth when learning the
ground-satellite image dictionary, whereas FAB-MAP does not use stereo.

\subsection{Malaga-Urban Dataset}
\begin{figure}[!t]
    \centering
    \includegraphics[width=0.95\linewidth]{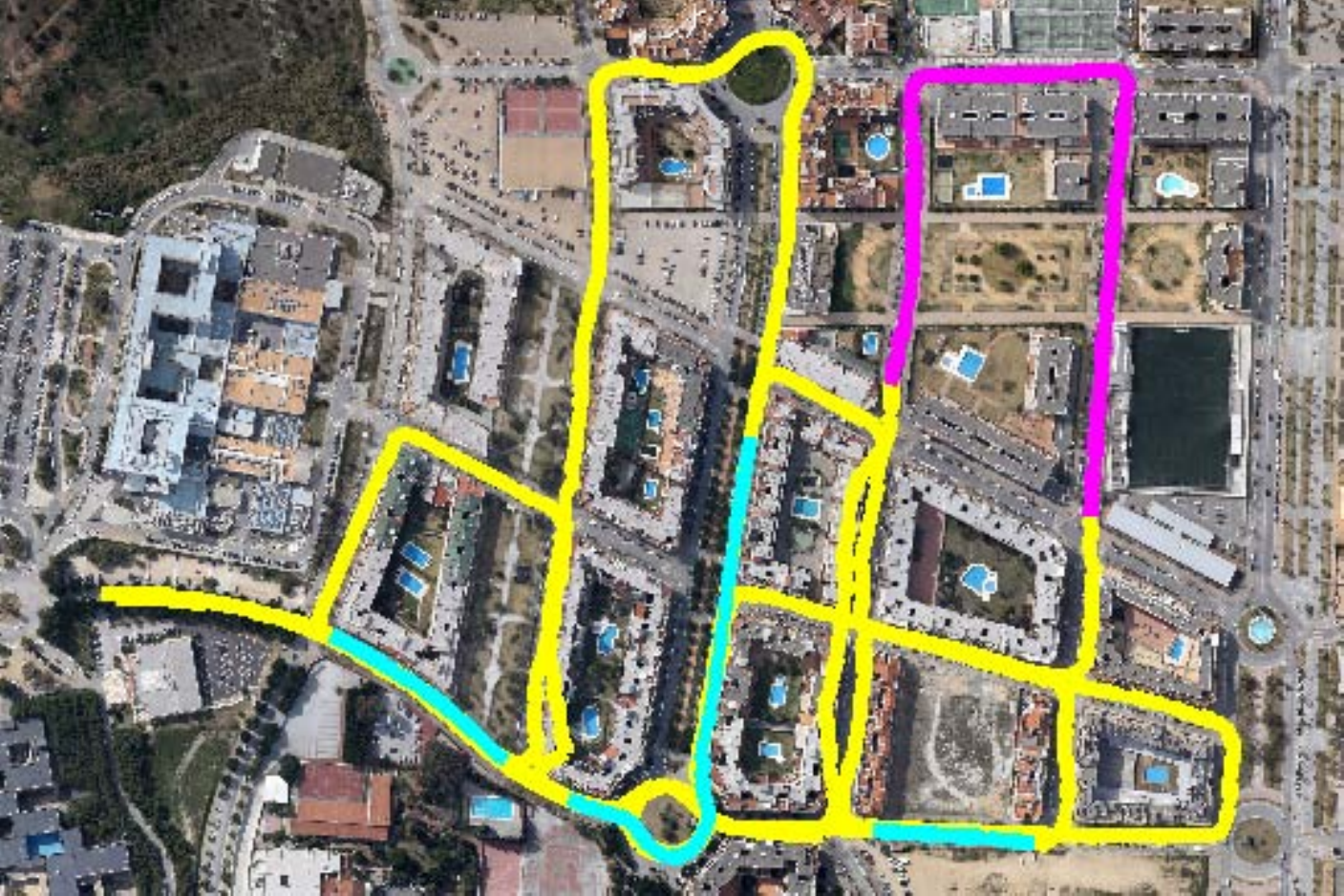}
    \caption{Data split for the Malaga sequence where yellow, cyan, and purple denote the database, revisit query, and the outside query set, respectively.}
    \label{fig:malaga}
\end{figure}
We also evaluate our framework on the Malaga-Urban
dataset~\cite{malaga}, where we adopt the setup similar to
KITTI-Residential, using the first vehicle pass of an area as the
database set, and the second visit as the query set. In addition, we
also set aside images taken from a path outside the area represented
in the
database to evaluate each method's ability to handle negative queries. We used the
longest sequence, Malaga-10, which contains $18203$ images, each with a
resolution of $1024\times 768$. We downsample the database and query
sets at approximately one frame per second. The total driving distance
is $6.08$\,km, with $4.96$\,km for the database set, $583.5$\,m as the inside
query set, and $534.3$\,m as the outside query set. Figure~\ref{fig:malaga}
depicts the data splits.

\begin{figure}[!t]
    \centering%
    \subfigure[GPS trajectory error]{\includegraphics[width=0.475\linewidth]{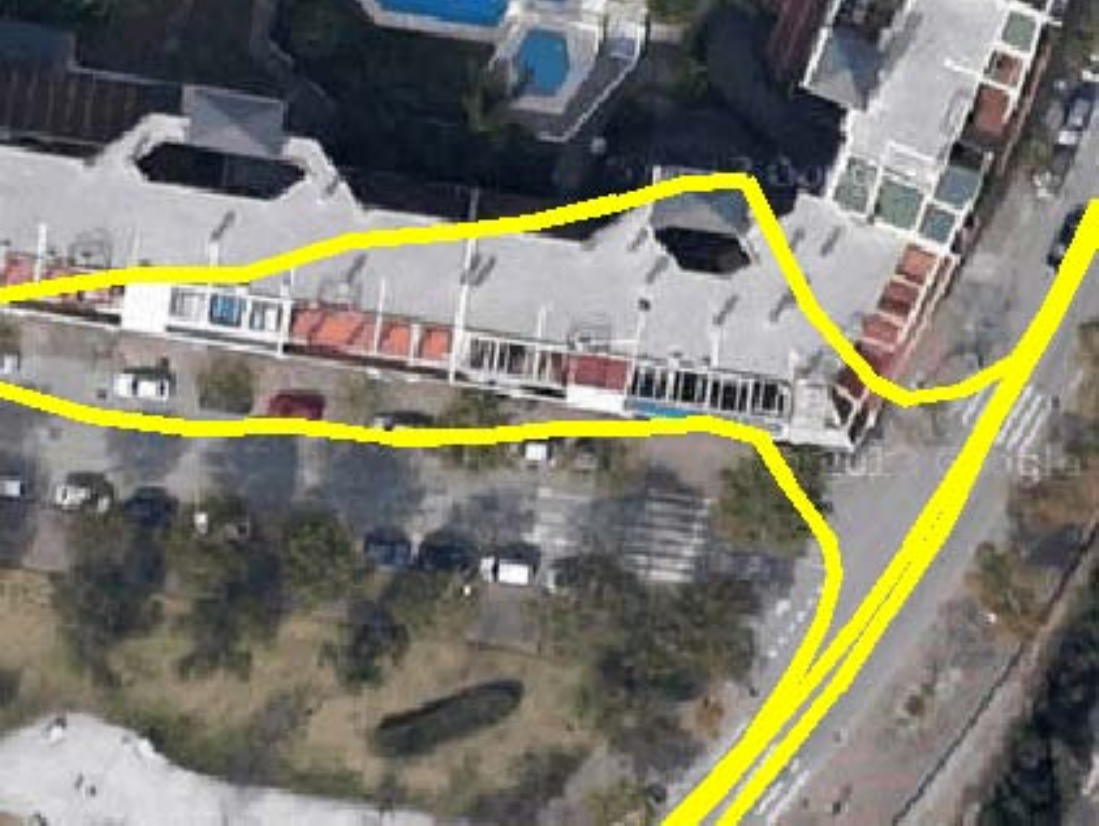}\label{fig:malaga-examples-trajectory}}\hfil
    \subfigure[Reference image]{\includegraphics[width=0.475\linewidth]{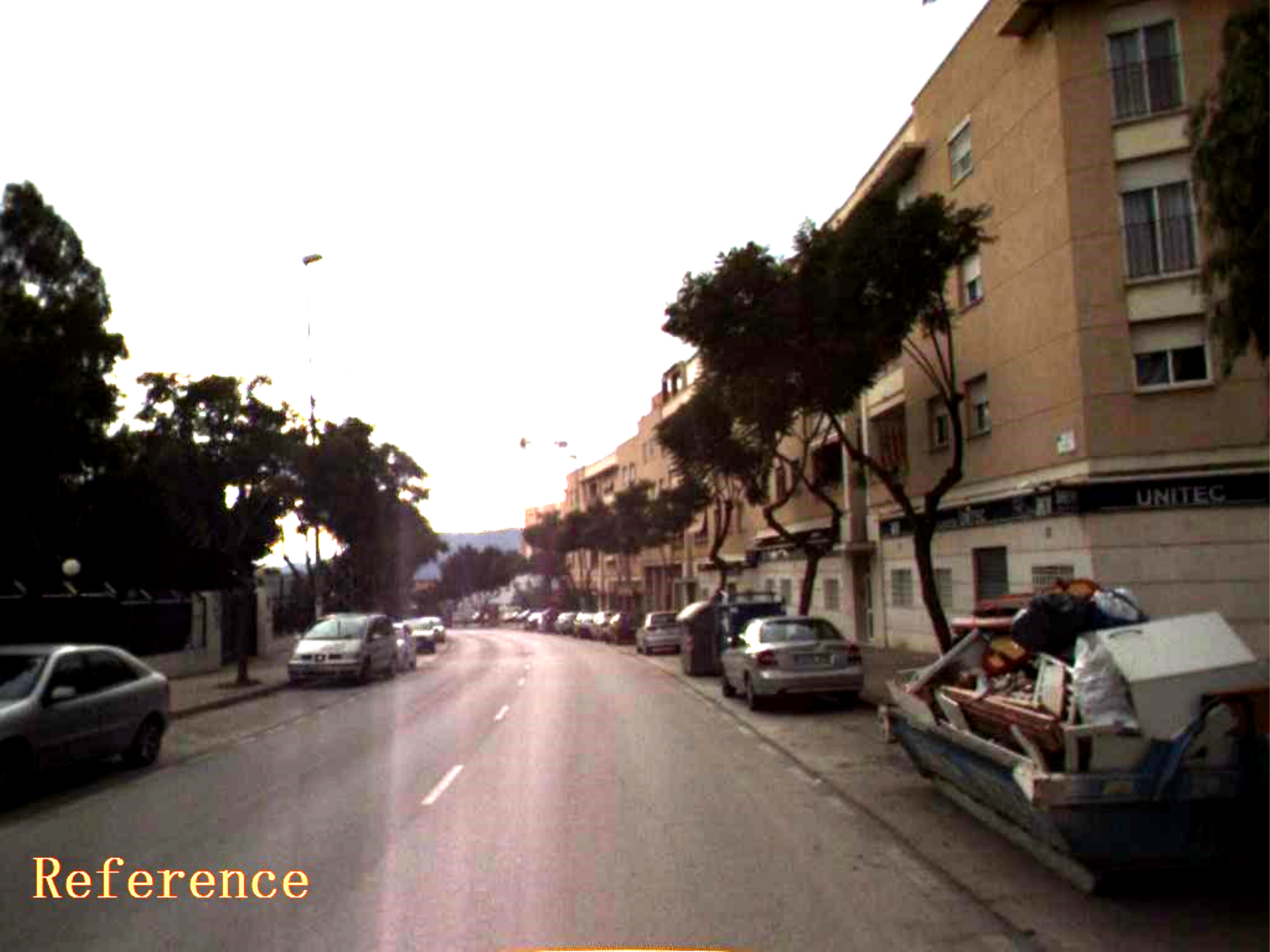}\label{fig:malaga-examples-reference}}\\
    \subfigure[GPS-based match]{\includegraphics[width=0.475\linewidth]{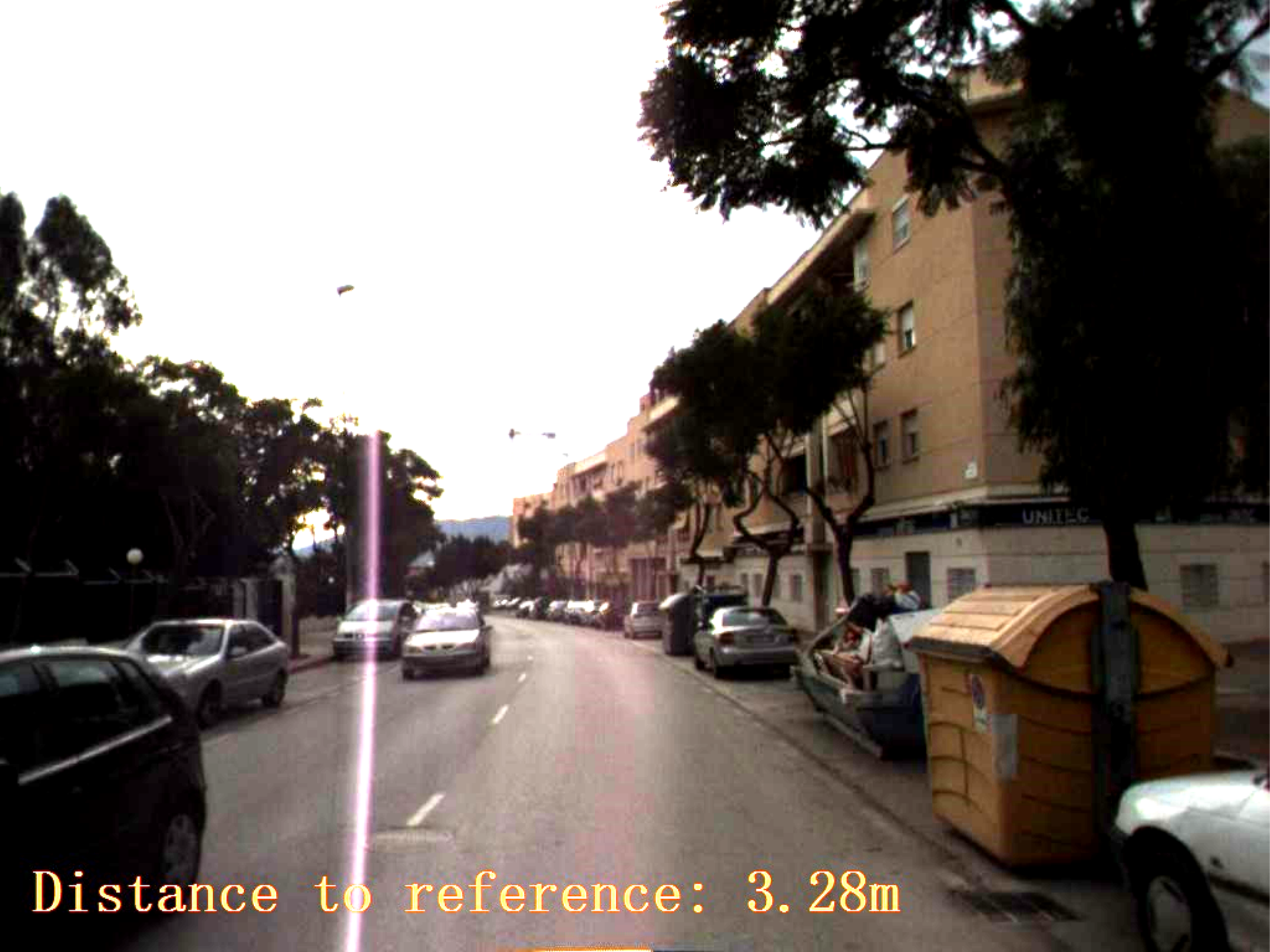}\label{fig:malaga-examples-gps}}\hfil
    \subfigure[Correct match]{\includegraphics[width=0.475\linewidth]{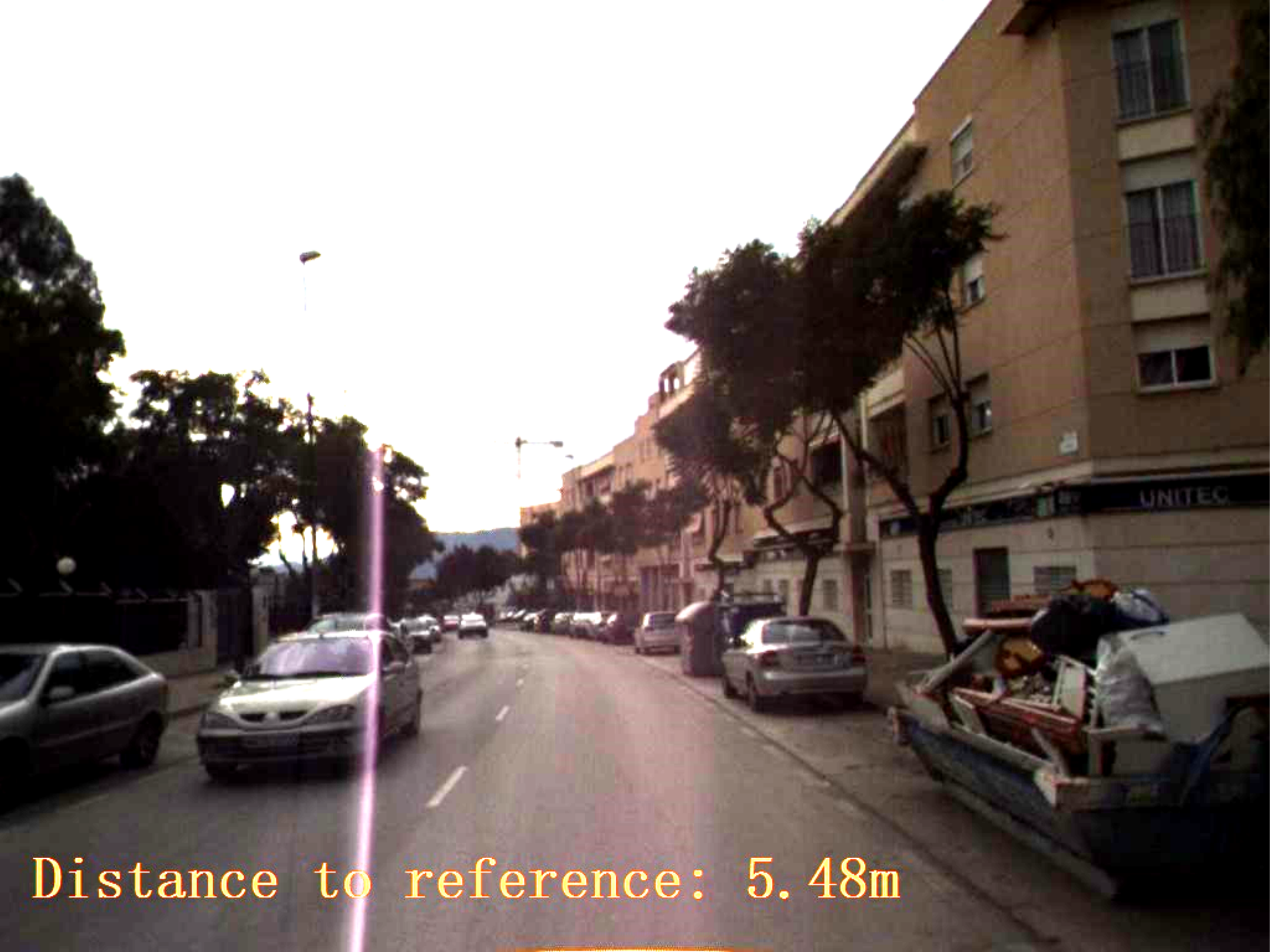}\label{fig:malaga-examples-correct}}%
    \caption{Deficiency of the ground-truth location tags in the Malaga
      dataset.}
\label{fig:malaga-examples}
\end{figure}
Unlike the KITTI datasets, the quality of the ground-truth location
tags in the Malaga dataset is relatively
poor. Figure~\ref{fig:malaga-examples} conveys the deficiencies in the
GPS tags provided with the data. Figure~\ref{fig:malaga-examples-trajectory} shows a portion of the
ground-truth trajectory, where the GPS data incorrectly suggests that
the vehicle took an infeasible path through the environment. As
further evidence, the GPS location tags suggest that the image in
Figure~\ref{fig:malaga-examples-gps} is $3.28$\,m away from the
reference image in Figure~\ref{fig:malaga-examples-reference} and constitutes
the nearest match. However, the image in Figure~\ref{fig:malaga-examples-correct} is
actually a closer match to the reference image (note the white trashcan in
the lower-right), despite being $5.48$\,m away according to the GPS
data. Due to the limited accuracy of the ground-truth locations, we
evaluate the methods in terms of place recognition (i.e., their ability to
localize the camera within $10$\,m of the ground-truth location) as
opposed to localization error.

\begin{figure}[!tb]
    \center
    \includegraphics[width=0.95\linewidth]{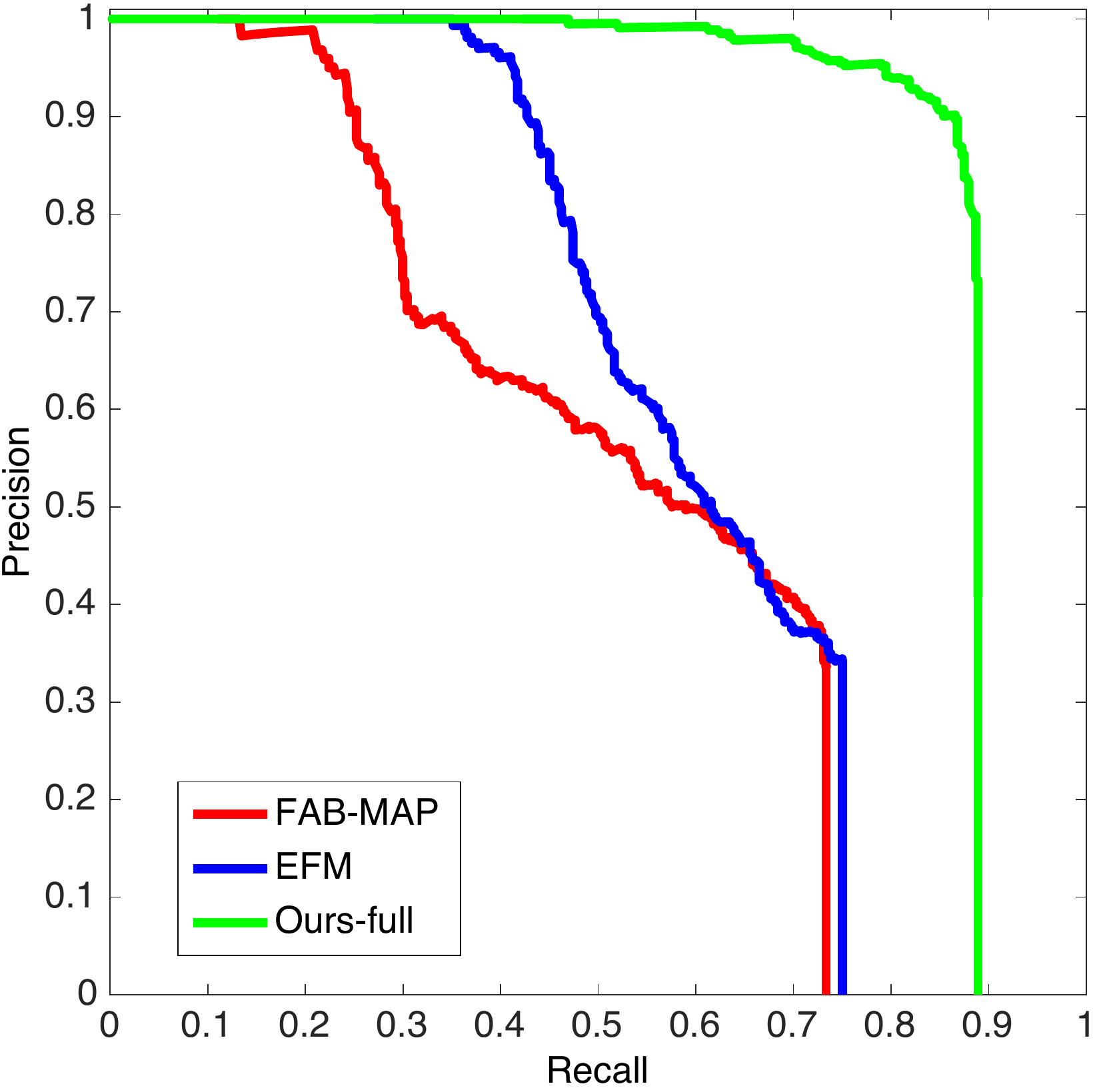}
    \caption{Precision-recall curve of different methods on the Malaga dataset.}
    \label{fig:malaga-precision-recall}
\end{figure}
We compare the full version of our algorithm to the FAB-MAP and Exhaustive
Feature Matching (EFM) methods as before. We set the bag-of-words size
for FAB-MAP to $2589$. We define true positives as images that are
identified as inliers and localized within $10$\,m of their
ground-truth locations. We picked optimal thresholds for each
method based on the square area under their precision-recall
curves (Fig.~\ref{fig:malaga-precision-recall}). Table~\ref{tab:malaga}
summarizes the precision and recall statistics for the different
methods.
\begin{table}[!t]
    \centering
    \caption{Optimal precision-recall of different methods.}
    \label{tab:malaga}
    \begin{tabularx}{0.55\linewidth}{lcc}
      \toprule
      Method                   & Precision                      & Recall            \\
      \midrule
      FAB-MAP \cite{cummins11}    & 49.3\%                            & 61.6\%               \\ 
      EFM                      & 89.4\%                            & 43.6\%            \\
      Ours-full                     & \textbf{90.2\%}                   & \textbf{86.6\%}            \\ 
      \bottomrule
\end{tabularx}
\end{table}

The results demonstrate that our method is able to correctly classify
most images as being inliers or outliers and subsequently estimate the
location of the inlier images. The EFM method
achieves comparable precision, however the computational
expense of doing exhaustive feature matching makes it intractable for
real-time use in all but trivially small environments. Note that using
only inlier images, the average location errors (with standard deviation) in
meters when rough localization succeeds for FAB-MAP, EFM, and our method
are $3.45 \, (2.16)$, $3.65 \, (2.21)$, and $3.33 \, (2.08)$, respectively. Although
our method achieves better accuracy, it is difficult to draw strong
conclusions due to the aforementioned deficiencies in the ground-truth
data.  We believe the improvement in accuracy of our method will
be more significant if accurate ground-truth location tags are
available, similar to what we have observed in our KITTI experiments.


\section{Conclusion} \label{sec:intro}

We presented a multimodal learning method that performs accurate
visual localization by exploiting the availability of satellite
imagery. Our approach takes a ground image as input and outputs the
vehicle's corresponding location on a georeferenced satellite image
using a learned ground-satellite image dictionary
embedding. We proposed an algorithm for
estimating the co-occurrence probabilities between the ground and
satellite images. We also described a ranking-based technique that
learns location-discriminative feature projection matrices, improving
the ability of our method to accurately localize a given
ground image. We evaluated our method on multiple public datasets,
which demonstrate its ability to accurately perform visual localization.

Our future work will focus on the inclusion of additional features
that enable learning with a smaller amount of data, and on learning
general ground-to-satellite relationships that generalize across
different environments.


\bibliographystyle{IEEEtranN}
\bibliography{references}

\end{document}